\documentclass{article}
\usepackage{spconf,amsmath,graphicx}
\usepackage{float}
\usepackage{multirow}
\usepackage{color,soul}
\usepackage{tabularx}
\usepackage{array}
\usepackage{physics}
\usepackage{pifont}
\usepackage{hyperref}
\hypersetup{
    colorlinks=true, 
    linkcolor=blue,  
    citecolor=blue,  
    urlcolor=blue,   
    filecolor=blue,  
    linktoc=all,     
    pdfnewwindow=true,
}

\title{Word-Level ASR Quality Estimation for Efficient Corpus Sampling and Post-Editing through Analyzing Attentions of a Reference-Free Metric}
\name{Golara Javadi, Kamer Ali Yuksel, Yunsu Kim, Thiago Castro Ferreira, Mohamed Al-Badrashiny}
\address{Author Affiliation(s)}
\address{aiXplain Inc., Los Gatos, CA, USA}
\begin{document}
%
\maketitle
\begin{abstract}
In the realm of automatic speech recognition (ASR), the quest for models that not only perform with high accuracy but also offer transparency in their decision-making processes is crucial. The potential of quality estimation (QE) metrics is introduced and evaluated as a novel tool to enhance explainable artificial intelligence (XAI) in ASR systems. Through experiments and analyses, the capabilities of the NoRefER (No Reference Error Rate) metric are explored in identifying word-level errors to aid post-editors in refining ASR hypotheses. The investigation also extends to the utility of NoRefER in the corpus-building process, demonstrating its effectiveness in augmenting datasets with insightful annotations.
The diagnostic aspects of NoRefER are examined, revealing its ability to provide valuable insights into model behaviors and decision patterns. This has proven beneficial for prioritizing hypotheses in post-editing workflows and fine-tuning ASR models.
The findings suggest that NoRefER is not merely a tool for error detection but also a comprehensive framework for enhancing ASR systems' transparency, efficiency, and effectiveness. To ensure the reproducibility of the results, all source codes of this study are made publicly available\footnote{\href{https://github.com/aixplain/NoRefER/tree/main/icassp-xai}{GitHub:https://github.com/aixplain/NoRefER/tree/main/icassp-xai}}.
\end{abstract}
\section{Introduction}
\label{sec:intro}
The quest for explainability in automatic speech recognition (ASR) systems has grown increasingly complex, mainly when dealing with commercial ASR models that often operate as ``black boxes". While delivering high accuracy, these complex models typically lack transparency in their decision-making processes. This lack of clarity poses significant challenges, especially in scenarios where understanding the rationale behind a model's decisions is equally as critical as the decisions themselves. Such scenarios include (but are not limited to) medical transcription, emergency response systems, accessibility services, and automotive systems, where the accurate processing of voice commands is essential for ensuring driver safety and vehicle functionality. 

\begin{figure}[t]
\centering 
\centerline{\includegraphics[width=\columnwidth]{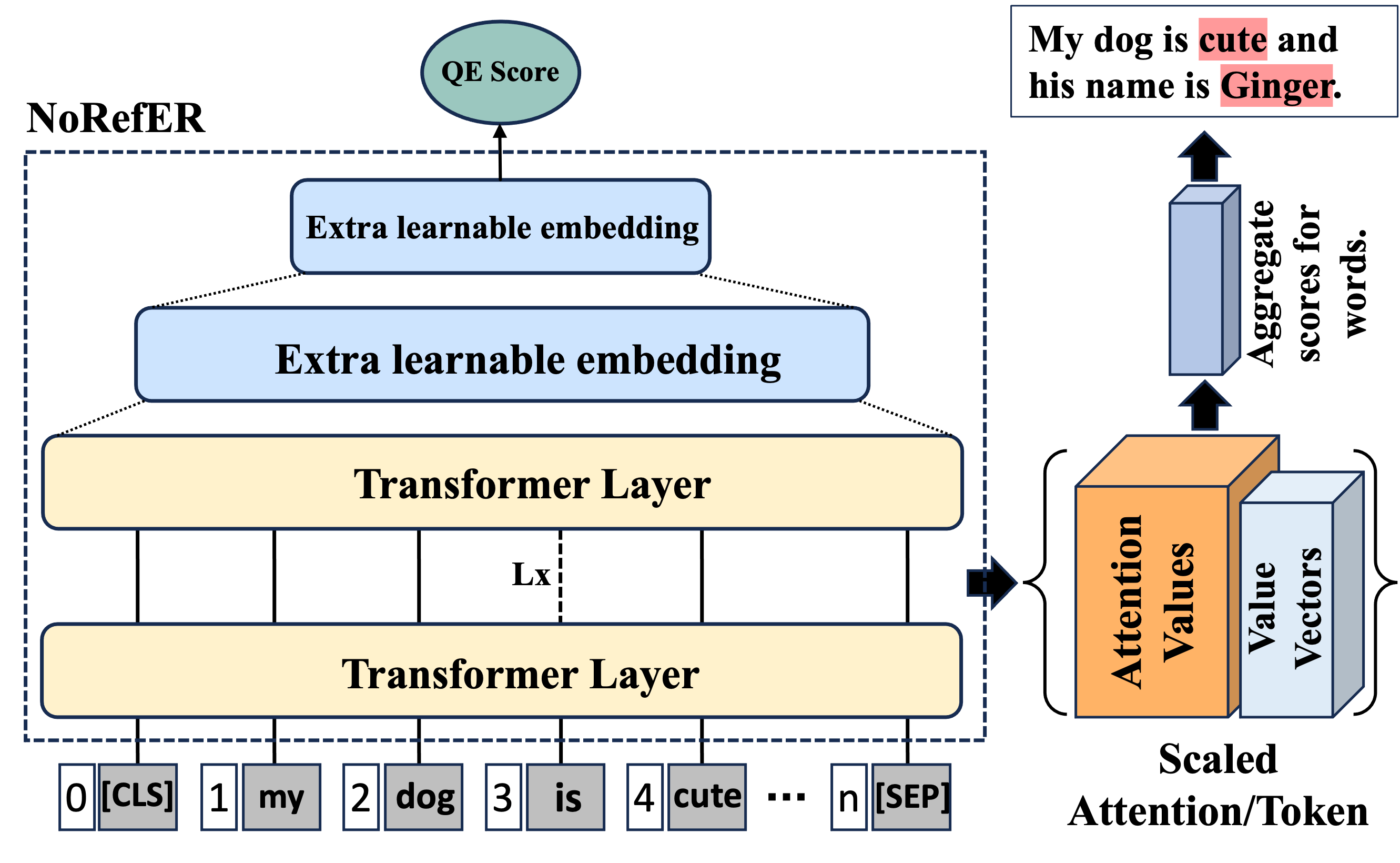}}
\caption{NoRefER attention for Error Identification in token and word level. Attention scores (with the L2 norm scaling) are averaged across all model layers for interpretability.}
\label{fig:method}
\end{figure}

Addressing this need, this paper explores the utility of token or word attention mechanisms of QE metrics for ASR explainability. The aim is to unravel the intricacies of these models, shedding light on how they process and interpret linguistic information. In recent years, attention mechanisms have become a cornerstone in the architecture of deep neural networks, especially in natural language processing (NLP) and ASR tasks. These mechanisms provide a means to interpret model decisions by assigning weights to different input parts. However, the direct interpretation of these attention weights is often misleading due to factors like entangled representations and the inherent instability of attention distributions \cite{voita2019analyzing}. An approach is employed in which attention weights are scaled with the norm of value vectors in transformer-based ASR models \cite{fomicheva-etal-2021-eval4nlp}. This method addresses raw attention weights' limitations, offering a more stable and reliable measure for identifying word-level errors (Figure~\ref{fig:method}).

The implications of this research extend to various aspects of ASR system deployment and improvement. By highlighting word-level errors through enhanced attention mechanisms, NoRefER can significantly aid post-editors and contribute to the corpus-building process for ASR systems. Traditional approaches to identifying inaccuracies in ASR outputs often require listening to entire audio files, which is time-consuming and labor-intensive. This method introduces a significant efficiency boost by pinpointing potential erroneous segments in the transcription. Such a targeted approach allows post-editors to focus their efforts directly on the problematic areas, dramatically reducing the time and effort involved in the correction process. \cite{yuksel2023efficient}
This study aims to bridge the gap between the advanced capabilities of ASR systems and the need for their explainability. The key contributions of this research are three-fold:
\begin{itemize}\itemsep0em
    \item NoRefER is introduced as a tool for enhancing ASR model interpretability by identifying word-level errors without reference transcriptions.
    \item NoRefER's role in assisting post-editors and improving ASR system quality through refined outputs and corpus-building is demonstrated.
    \item The application of NoRefER to commercial (black-box) models is showcased, enabling error identification and enhancing transparency without confidence scores.
\end{itemize}

\section{Related works}
\label{sec:related}

QE in ASR is identified as a critical aspect of evaluating and enhancing the performance of ASR systems without references. Traditionally, evaluation has been centered around reference-based metrics --like Word Error Rate (WER)-- which compare ASR outputs with ground-truth transcripts. However, obtaining these transcripts is often time-consuming and costly. A novel approach to QE is the reference-less metrics \cite{rei-etal-2020-unbabels,yuksel2023industry,chowdhury2023multilingual,yuksel23_icassp}.
One of these metrics is NoRefER, a multi-language reference-free quality metric for ASR systems \cite{yuksel23_icassp,yuksel23_interspeech}. The innovation in NoRefER lies in its ability to provide an evaluation metric that overcomes the limitations of traditional reference-based metrics, enabling its application to speech datasets lacking ground truth. The fine-tuned model demonstrated a high correlation with WER scores and ranks.

In ASR and language technologies, explainability becomes crucial for identifying the reasons behind a model's predictions, enhancing transparency, and building trust. Explainability methods can be broadly categorized into simplification-based, perturbation-based, and gradient-based approaches. One key approach is feature attribution methods, which assign relevance scores to inputs to explain model predictions. These methods include techniques like local interpretable model-agnostic explanations (LIME) \cite{ribeiro2016should}, gradients, and attention mechanisms \cite{treviso2021unbabel,kobayashi2020attention}. An exciting application of explainability in ASR is observed in QE, where feature attribution methods are employed to understand how QE models make predictions. \cite{fomicheva2021translation} discusses how feature attribution methods can derive information on translation errors from sentence-level models in QE. 
\cite{voita2019bottom} also highlights the role of layer-wise token representation in Transformers, which is key to using NoRefER attention for explainability. 

\section{Methodology}
\label{sec:method}
In this work, the attention distribution of the NoRefER model is closely examined, focusing mainly on its behavior at both word and token levels. This analysis targets sentences with varying WER. The aim is to identify critical patterns or tendencies that could enhance the model's performance and inform more effective training strategies. The NoRefER architecture is centered around a pre-trained cross-lingual language model (LM), MiniLMv2, a compact and efficient XLM-RoBERTa-Large \cite{wang2020minilmv2} derivative containing only 117 million parameters with the addition of two linear layers, a dropout, and a non-linear activation function in-between \cite{yuksel23_icassp}.

Initially, attention scores from the NoRefER model’s layers of the encoder before the linear layers are collected. These scores represent the degree of focus each token places on other tokens within a sentence batch. Structured as tensors, these scores require aggregation for interpretation in ASR-QE. First, attention scores are averaged across all layers and heads \cite{voita2019bottom}.
Then, the token-level attention is calculated by averaging scores directed by each token toward others, assessing their contextual significance. Lastly, the interpretability of attention values is further refined by scaling the attention probabilities with the L2 norm of value vectors. This approach, reflecting the strategies in \cite{fomicheva-etal-2021-eval4nlp}, merges the magnitude of value vectors with attention probabilities, improving the performance of word-level models. Such scaling is demonstrated to yield superior results in experiments, surpassing other methods in accuracy and correlation metrics.

In transitioning from token-level to word-level attention within the analysis, max-pooling is adopted to aggregate the attention scores effectively.
This pinpoints the most impactful token within a word, thereby reflecting the word's overall significance or emphasis in the model's attention mechanism.
Following the aggregation of attention scores, the next step involves a detailed investigation of the relationship between these attention scores and the actual transcription errors. This analysis aims to ascertain whether the attention mechanism, as interpreted through the method, aligns with the occurrence of errors in transcriptions. Insights into the model's error detection capabilities are intended to be unveiled by correlating attention scores with error patterns.

The study sets to compare the effectiveness of the NoRefER's attention-based error detection with the conventional approach of using confidence scores from open-source ASR models, such as CTC and Whisper \cite{gulati2020conformer,radford2023robust}. This comparison is particularly pertinent in cases where ASR confidence scores are unavailable, highlighting the NoRefER's utility in practical, real-world ASR applications.

\section{Dataset and metrics}
\label{sec:experiments}

For evaluation, the referenceless metric, NoRefER, is applied to Common Voice (in English, French, Spanish) \cite{ardila2019common} and LibriSpeech (English) \cite{panayotov2015librispeech} test datasets. Transcription hypotheses for these test datasets are sourced from leading commercial ASR engines (AWS, AppTek, Azure, Deepgram, Google, and OpenAI's Whisper-Large), ensuring a thorough evaluation of each speech segment within the datasets. The median sentence length and the median of faulty words per sentence in each dataset are as follows: 

\begin{table}[ht]
\centering
\caption{Sentence Length and Error Statistics in Datasets}
\label{tab:dataset-stats}
\setlength{\tabcolsep}{3pt}
\begin{tabular}{lcc}
\hline
\textbf{Dataset} & \textbf{Median length} & \textbf{Median errors/sent} \\ \hline
LibriSpeech (En)         & 17 & 3 \\ 
Common Voice        & 10 & 1 \\ 
All Datasets         & 11 & 2 \\ \hline
\end{tabular}
\end{table}

In assessing the performance of the NoRefER metric's attention mechanism, the below suite of metrics is applied, designed to evaluate both continuous and binary scoring methods. These metrics are calculated for each instance in the test set and then averaged. 
This approach aligns with established practices in \cite{atanasova2020diagnostic} for robustly assessing word-level outputs.

\textbf{Area Under the Curve (AUC)}: Each transcription instance is evaluated by computing the AUC score, which measures the model's ability to distinguish between correctly transcribed words and errors based on continuous attention scores against binary error labels from ground-truth references.

\textbf{Average Precision (AP)}: Recognizing the limitations of AUC in the context of imbalanced datasets, Average Precision is also employed as a metric. AP provides a more detailed assessment by summarizing the precision-recall curve through the weighted mean of precision at successive thresholds, with weights derived from the increase in recall.

\textbf{Top-K classification metrics}: These metrics (Precision, Recall, F1, Accuracy), commonly utilized in ranking, calculate the fraction of words with the highest attention scores corresponding to actual transcription errors relative to the total number of errors present in the output. They measure the capability of NoRefER to prioritize words for post-editing or further scrutiny. Here, the weighted versions of those metrics are always employed to account for the class imbalance.

\begin{table*}[ht]
\centering
\caption{The comparative analysis of attention impact on word-level error-ranking metrics using the NoRefER metric, and comparison of the best result with several baseline methods, incorporating gradients or confidence, as an alternative to attention.}
\label{tab:k1}
\begin{tabular}{l|c|ccc|cc|c|c} 
\hline
\textbf{Dataset} & \textbf{Variant} & \begin{tabular}[c]{@{}c@{}}\textbf{Recall}\\\textbf{@k}\end{tabular} & \begin{tabular}[c]{@{}c@{}}\textbf{Precision}\\\textbf{@k}\end{tabular}  & \begin{tabular}[c]{@{}c@{}}\textbf{F1}\\\textbf{@k}\end{tabular} & \begin{tabular}[c]{@{}c@{}}\textbf{Accuracy}\\\textbf{@k}\end{tabular} & \begin{tabular}[c]{@{}c@{}}\textbf{Balanced}\\\textbf{Acc @k}\end{tabular} & \textbf{AP} & \textbf{AUC}\\ 
\hline 
\multirow{2}{*}{en-librispeech}         
    & $k=2$ & 0.80 & 0.81 & 0.79 & 0.80 & 0.66 & \multirow{2}{*}{0.62} & \multirow{2}{*}{0.78} \\
    & dyn & 0.81 & 0.82 & 0.80 & 0.81 & 0.68 & & \\
\multirow{2}{*}{en-common}        
    & $k=2$ & 0.71 & 0.81 & 0.74 & 0.71 & 0.63 & \multirow{2}{*}{0.36} & \multirow{2}{*}{0.63}\\
    & dyn & 0.75 & 0.82 & 0.76 & 0.75 & 0.65 & & \\
\multirow{2}{*}{es-common}        
    & $k=2$ & 0.71 & 0.81 & 0.73 & 0.71 & 0.64 & \multirow{2}{*}{0.38} & \multirow{2}{*}{0.66}\\
    & dyn & 0.75 & 0.81 & 0.76 & 0.75 & 0.66 & & \\
\multirow{2}{*}{fr-common}        
    & $k=2$ & 0.65 & 0.72 & 0.65 & 0.65 & 0.59 & \multirow{2}{*}{0.57} & \multirow{2}{*}{0.68}\\
    & dyn & 0.68 & 0.75 & 0.66 & 0.68 & 0.61 & & \\  \hline
\multirow{2}{*}{all-datasets (proposed)}     
    & $k=2$ & 0.72 & 0.79 & 0.73 & 0.72 & 0.63 & \multirow{2}{*}{\textbf{0.48}} & \multirow{2}{*}{\textbf{0.69}}\\
    & dyn & \textbf{0.75} & \textbf{0.80} & \textbf{0.75} & \textbf{0.75} & \textbf{0.65} & & \\ 
    & gradient & 0.72 & 0.77 & 0.72 & 0.72 & 0.60 & 0.31 & 0.62 \\ \hline

\multirow{2}{*}{all-datasets (baselines)}  
    & whisper & 0.65 & 0.73 & 0.64 & 0.65 & 0.56 & 0.49 & 0.62\\
    & ctc-conf & 0.52 & 0.91 & 0.57 & 0.52 & 0.51 & 0.52 & 0.74\\     
    & random & 0.12 & 0.23 & 0.14 & 0.70 & 0.56 & 0.31 & 0.50 \\ 
    \hline
\end{tabular}
\end{table*}

\section{Experiments}
\label{sec:exp}
Using max-pooling and attention scaled by the norm of value vectors as the chosen method for aggregating attention scores, the metrics outlined in Section \ref{sec:experiments} are calculated to evaluate the attention scores as a QE measure. As a reference, attention values are compared with the binary scores obtained using the JiWER library \cite{jiwer2023}, which labels faulty words in the transcription, such as those that are deleted, inserted, or substituted. This comparison is crucial in assessing the effectiveness of the attention scores in identifying errors within referenceless transcription hypotheses. As a baseline for this evaluation, confidence scores from publicly available ASR models, including Whisper and CTC, are utilized (Table~\ref{tab:k1}).

\begin{figure}[ht]
\centering 
\centerline{\includegraphics[width=.74\columnwidth]{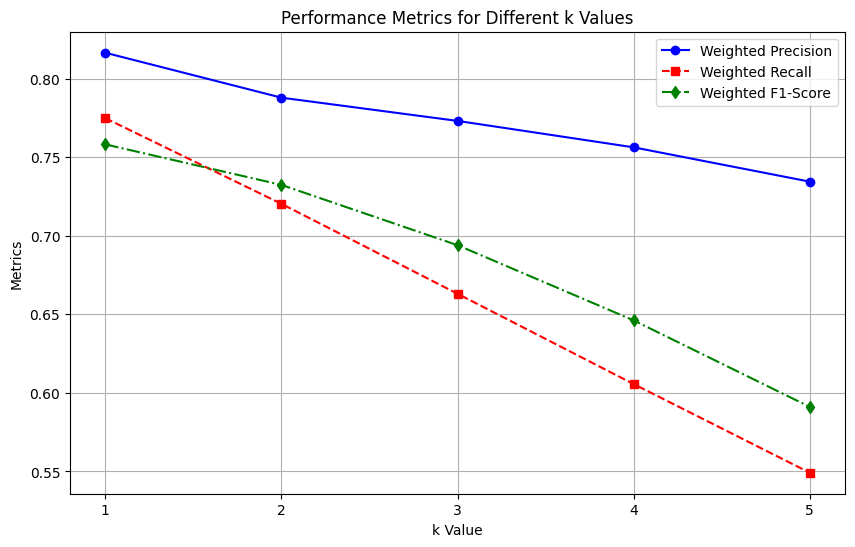}}
\caption{Variation of Weighted Precision, Recall, and F1-Score with increasing $k$ in the evaluation of word-level attention.}
\label{fig:k_selection}
\end{figure}

Two distinct approaches are investigated when selecting the optimal value for $k$ in the classification metrics. The first approach involves fixing $k$ to a constant value of 1 to 5. This range is visually explored through a plot, as seen in Figure~\ref{fig:k_selection}, showing performance metrics across different $k$ values. Since the median number of faulty words per sentence in the datasets is 2, setting $k$ to 2 is identified as a logical choice. Furthermore, analysis reveals that the metrics at $k=1$ yield promising results, suggesting that even the most highly weighted words in a sentence offer significant information about potential errors. The second method adopts a dynamic approach, setting $k$ to 10 percent of the length of each sentence. This adaptive strategy considers the variability in sentence lengths across the dataset, thus providing a tailored evaluation for each transcription. The results of this dynamic $k$ selection method are detailed in Table \ref{tab:k1}, demonstrating its robustness in assessing the ASR quality.

For the ablation study, two strategies for explainability are explored to understand the model's behavior comprehensively. These include using raw attention values, attention scaled by the norm of value vectors, and multiplying with gradients. By scaling attention with the norm of value vectors, deeper insights are gained into how the model weights and prioritizes different tokens and words in a sentence.
In transitioning from token-level to word-level attention, three aggregation methods are examined: Average Attention, which calculates the mean of all constituent token scores for each word; Max Attention, identifying the highest attention score within a word's tokens; and Q3 Attention, focusing on the third quartile value among token scores to mitigate the impact of outliers. Max Attention is ultimately selected for its efficacy in pinpointing the most influential parts of a word according to the model’s attention. Table \ref{tab:ablation_study} compares all settings, demonstrating the rationale behind choosing Max Attention and scaling by the norm of value vectors.

\begin{table}[t]
\centering
\caption{Ablation studies on attention aggregation and normalization methods for the NoRefER metric over all datasets.}
\label{tab:ablation_study}
\begin{tabular}{lcc} \hline
\textbf{Aggregation} & \textbf{Scaling} & \textbf{F1@k} \\ \hline
avg & $att * value\,norm$ & 0.70 \\
q3 & $att * value\,norm$ & 0.73 \\
max & $att * value\,norm$ & \textbf{0.75} \\
max & $att * norm * grad$ & 0.70 \\
max & $raw\;attention\;(att)$ & 0.70 \\ \hline
\end{tabular}
\end{table}

\section{Discussion}
\label{sec:discussion}

Remarkably, the findings reveal that the attention values generated by the NoRefER model are adept at distinguishing faulty words. This is significant, as it demonstrates the metric's capability to identify potential errors in the transcriptions without needing reference transcriptions. These results are found to be even superior to those obtained using traditional ASR confidence scores. This superiority not only reinforces the effectiveness of the NoRefER model in error detection but also underscores its advantage in scenarios where ASR confidence scores are either unavailable or unreliable. 

In the dataset-level analysis, words are ranked based on the frequency of their erroneous occurrences across the entire dataset. The average attention score is calculated for each word, aggregating across all instances of its appearance. This process allows for an evaluation of how the model's attention correlates with the rate of errors associated with each word throughout the dataset. The high correlation between attention scores and word error frequencies (Table \ref{tab:datasetcor}) highlights the corpus-building potential of the NoRefER model. The ability to identify and quantify error-prone words based on attention provides valuable insights and guidelines for potential adjustments in model training and data augmentation.

To actualize this, data augmentation can be strategically implemented by introducing variations of the identified error-prone words in diverse linguistic contexts. This process enriches the training dataset, allowing the ASR system to learn from a broader spectrum of scenarios where these specific errors may occur. Additionally, targeted post-editing and fine-tuning can be applied, wherein the model is repeatedly exposed to and corrected for these particular errors. This iterative process sharpens the model's understanding of language patterns and reinforces its ability to interpret and transcribe speech accurately in complex or ambiguous situations. 

\begin{table}[t]
\centering
\caption{Correlation calculated between NoRefER attention scores and actual word error frequencies in various datasets.}
\label{tab:datasetcor}
\begin{tabular}{lccc}
\hline
\textbf{Dataset} & \textbf{Pearson} & \textbf{Kendall} & \textbf{Spearman} \\ \hline
en-librispeech & 0.87 & 0.68 & 0.82 \\
en-common & 0.87 & 0.68 & 0.82 \\
es-common & 0.91 & 0.73 & 0.86 \\
fr-common & 0.74 & 0.60 & 0.74 \\ \hline
all-datasets & 0.80 & 0.65 & 0.79 \\ \hline
\end{tabular}
\end{table}

\section{Conclusions}
\label{sec:conclusion}
In conclusion, the NoRefER tool significantly advances XAI in ASR systems, excelling in identifying word-level errors and refining ASR outputs. Furthermore, its dataset-level analysis, correlating attention scores with error rates, showcases its potential to enhance model training and data augmentation strategies. The findings establish NoRefER as a robust tool for ASR error detection, as well as interpretability and efficacy; marking a notable advancement in the field of XAI.

\newpage
\bibliographystyle{IEEEbib}
\bibliography{main}

\end{document}